\documentclass[11pt,a4paper]{article}
\usepackage{scrextend}
\usepackage{hyperref}
\usepackage[hyperref]{emnlp2020}
\usepackage{times}
\usepackage{latexsym}
\usepackage{color,soul}
\setul{0.45ex}{0.25ex}
\usepackage{graphicx}
\usepackage{subcaption}
\usepackage{booktabs}
\usepackage{multirow, makecell}
\usepackage{tikz}
\usepackage{graphicx}
\usepackage{array}
\usepackage{pgfplots}
\usepackage{algorithm}
\usepackage{textcomp}
\usepackage{xcolor}
\usepackage{times}
\usepackage{latexsym}
\usepackage{amsmath}
\usepackage{amsfonts}
\usepackage[noend]{algpseudocode}
\usepackage{lipsum}
\usepackage{multirow}
\usepackage{framed}
\usepackage{graphicx}
\usepackage{longtable}
\usepackage{tikz}
\usepackage{booktabs}
\usepackage[all]{nowidow}
\usepackage{enumitem}
\usepackage{color,soul}
\usepackage{amssymb}
\usepackage{bm}
\usepackage{cleveref}
\crefformat{section}{\S#2#1#3} 
\crefformat{subsection}{\S#2#1#3}
\crefformat{subsubsection}{\S#2#1#3}
\definecolor{object_orange}{rgb}{1,0.4,0}

\newcommand*{\txtfont}{\fontfamily{txtt}\selectfont}
\newcommand{\txtstyle}[1]{{\small\txtfont #1}}
\newcommand{\txtstyless}[1]{{\scriptsize\txtfont #1}}
\newcommand{\objstyle}[1]{\textcolor{object_orange}{\textbf{#1}}}

\usepackage{microtype}

\aclfinalcopy 

\newif\ifcomments
\commentstrue 
\ifcomments
    \newcommand\gamaga[1]{[\textcolor{teal}{GI: {#1}}]}
    \newcommand\ali[1]{[\textcolor{olive}{AF: {#1}}]}
    \newcommand\hanna[1]{[\textcolor{purple}{HH: {#1}}]}
    \newcommand\rowan[1]{[\textcolor{magenta}{RZ: {#1}}]}
\else
    \providecommand{\gamaga}[1]{}
    \providecommand{\ali}[1]{}
    \providecommand{\hanna}[1]{}
    \providecommand{\rowan}[1]{}
\fi

\title{Probing Contextual Language Models for Common Ground\\with Visual Representations}

\date{}

\author{
Gabriel Ilharco \quad Rowan Zellers \quad  Ali Farhadi \quad Hannaneh Hajishirzi \\
  Paul G. Allen School of Computer Science \& Engineering\\ 
  University of Washington \\
  {\tt \{gamaga,rowanz,ali,hannaneh\}@cs.washington.edu}\\
}

\date{}

\begin{document}
\maketitle
\begin{abstract}

The success of large-scale contextual language models has attracted great interest in probing what is encoded in their representations.
In this work, we consider a new question: to what extent contextual representations of concrete nouns are aligned with corresponding visual representations? 
We design a probing model that evaluates how effective are text-only representations in distinguishing between matching and non-matching visual representations. Our findings show that language representations alone provide a strong signal for retrieving image patches from the correct object categories.
Moreover, they are effective in retrieving specific instances of image patches; textual context plays an important role in this process.
Visually grounded language models slightly outperform text-only language models in instance retrieval, but greatly under-perform humans.
We hope our analyses inspire future research in understanding and improving the visual capabilities of language models.

\end{abstract}

\section{Introduction}
\label{sec:introduction}

\renewcommand{\floatpagefraction}{0.99}

Contextual language models trained on text-only corpora are prevalent in recent natural language processing (NLP) literature \cite{devlin-etal-2019-bert, liu2019roberta, lan2019albert, 2019t5}. Understanding what their representations encode has been the goal of a number of recent studies \cite{belinkov2019analysis,rogers2020primer}. Yet, much is left to be understood about whether---or to what extent---these models can encode visual information.

We study this problem in the context of language grounding \cite{searle1984minds, harnad1990symbol, mcclelland2019extending, bisk2020experience, bender-2020-climbing}, empirically investigating  whether text-only representations can naturally be connected to the visual domain, without explicit visual supervision in pre-training.

\begin{figure}
  \centering   
    \includegraphics[width=.99\linewidth,clip]{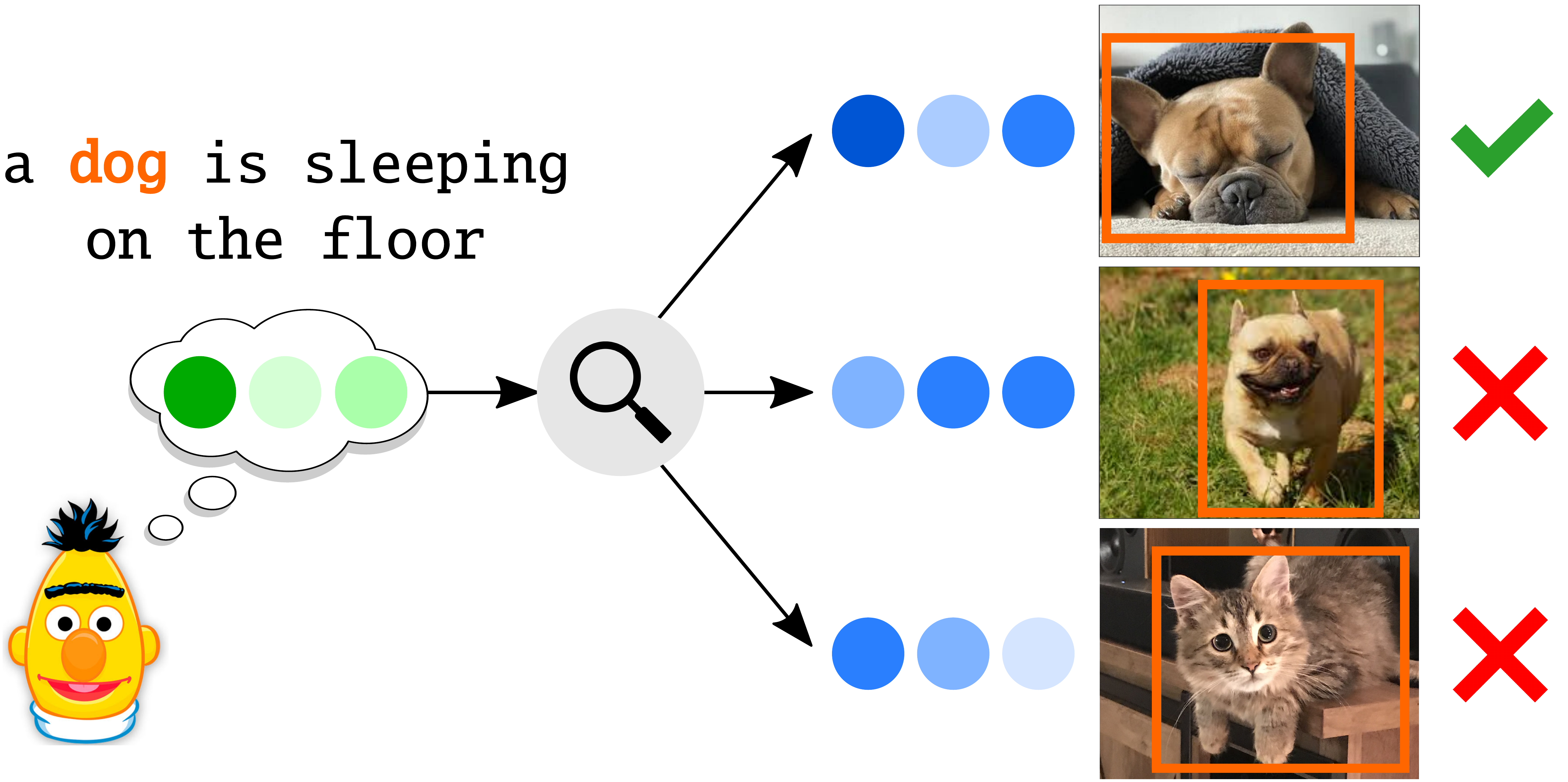}
  \caption{
  We introduce a probing mechanism that learns a mapping from contextual language representations to visual features. For a number of contextual language models, we evaluate how useful their representations are for retrieving matching image patches.}
\label{fig:teaser}
\end{figure}

We argue that {\it context} plays a significant role in this investigation.
In language, the ability to form context-dependent representations has shown to be crucial in designing pre-trained language models \cite{peters2018deep, devlin-etal-2019-bert}.
This is even more important for studying grounding since many visual properties depend strongly on context \cite{visualphrases}. For instance, a ``\txtstyle{flying \objstyle{bat}}" shares very few visual similarities with a ``\txtstyle{baseball \objstyle{bat}}"; likewise, a ``\txtstyle{\objstyle{dog} sleeping}" looks different from a ``\txtstyle{\objstyle{dog} running}". While alignments between language representations and visual attributes have attracted past interest \cite{leong2011measuring,lazaridou-etal-2014-wampimuk, lazaridou-etal-2015-combining,lucy-gauthier-2017-distributional,collell2017imagined}, the role of context has been previously overlooked, leaving many open questions about what visual information contextual language representations encode.

\begin{figure*}
  \centering   
\includegraphics[width=.89\linewidth,clip]{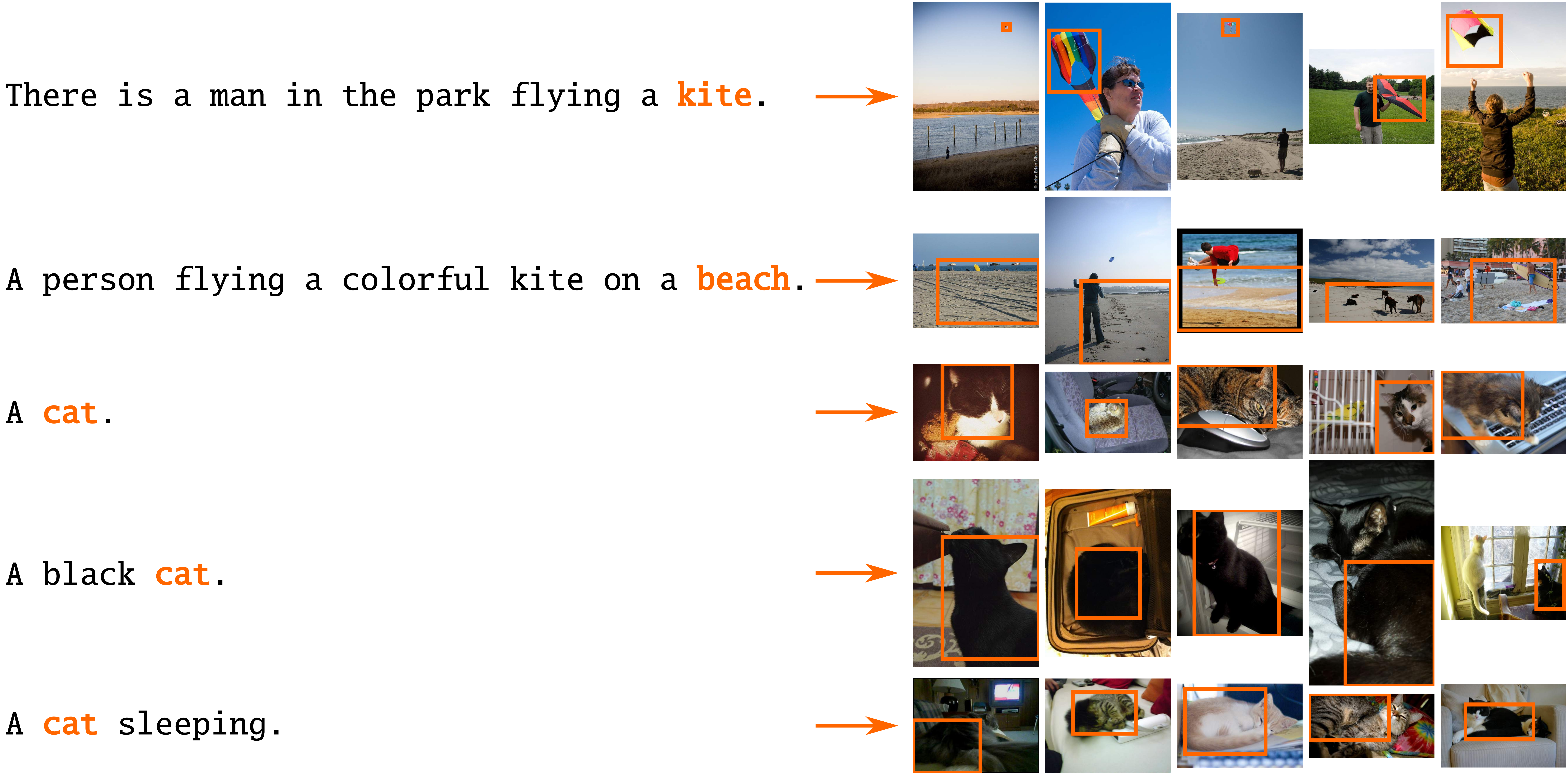}
  \caption{Examples of retrieved image patches from text-only representations using our probe. All shown images are retrieved from MS-COCO \cite{lin2014microsoft}, using representations from BERT base. Importantly, these object categories (e.g. \txtstyle{\objstyle{kite}}) are previously \textit{unseen} by our probe. On the bottom rows, we show examples of the influence of context in retrieval: while all retrieved image patches belong to the correct object category, \txtstyle{\objstyle{cat}}, more descriptive contexts allow more accurate retrieval at the instance level.}
\label{fig:examples}
\end{figure*}

In this work, we introduce a method for empirically {\it probing} contextual language representations and their relation to the visual domain. In general, probing examines properties for which the models are not designed to predict, but can be encoded in their representations  \cite{shi-etal-2016-string, rogers2020primer}. 
Here, our probe is a lightweight model trained to map language representations of concrete objects to  corresponding visual representations. The probe  (illustrated in Figure \ref{fig:teaser})  measures whether language representations can be used to give higher scores to matching visual representations compared to mismatched ones. 

Textual and visual representations are collected from image captioning data, where we find pairs of concrete words (e.g. \txtstyle{\objstyle{cat}} or \txtstyle{\objstyle{kite}}) and their corresponding image patches.
The probe is trained  using a contrastive loss \cite{oord2018representation} that gauges the mutual information between the language and visual representations. 
Given text-only representations of an unseen object category, the trained probe is evaluated by retrieving corresponding image patches for categories it has never seen during training. Qualitative examples can be found in Figure~\ref{fig:examples}.

We  examine representations from a number of contextual language models including BERT, RoBERTa, ALBERT and T5 \cite{ devlin-etal-2019-bert, liu2019roberta, lan2019albert, 2019t5}.
For all of them, we find that interesting mappings can be learned from language to visual representations, as illustrated in Figure \ref{fig:examples}. In particular, using its top-5 predictions, BERT representations retrieve the correctly paired visual instance 36\% of the time, strongly outperforming non-contextual language models (e.g., GloVe \cite{pennington2014glove}). Moreover, for all examined models, image patches of the correct object category are retrieved with a recall of 84-90\%. Our experiments are backed by a control task where visual representations are intentionally mismatched with their textual counterparts. Retrieval performance drops substantially in these settings, attesting the selectivity of our probe.

Moreover, we measure the impact of context on retrieval at the instance level. Contextual models substantially outperform non-contextual embeddings, but this difference disappears as context is gradually hidden from contextual models. When the context includes adjectives directly associated with the noun being inspected, we find significantly better instance retrieval performance.

Finally, we investigate a number of grounded language models---such as LXMERT and VILBERT \cite{tan-bansal-2019-lxmert,lu2019vilbert,lu201912}---that see visual data in training, finding them to slightly outperform text-only models. 
Contrasting the learned mappings with human judgment, the examined visually grounded language models significantly underperform human subjects, exposing much room for future improvement.

\section{Related Work}
\label{sec:related_work}

\paragraph{What is encoded in language representations?} Understanding what information NLP models encode has attracted great interest in recent years \cite{rogers2020primer}. From factual \cite{petroni-etal-2019-language, jawahar2019does, roberts2020much} to linguistic \cite{conneau2018you, liu2019linguistic, talmor2019olmpics} and commonsense \cite{forbes2019neural} knowledge, a wide set of properties have been previously analysed.  We refer to \citet{belinkov2019analysis} and \citet{rogers2020primer} for a more comprehensive literature review. A common approach, often used for inspecting contextual models, is probing \cite{shi-etal-2016-string, adi2016fine, conneau2018you, hewitt2019control}. In short, it consists of using supervised models to predict properties not directly inferred by the models. Probing is typically used in settings were discrete, linguistic annotations such as parts of speech are available. Our approach differs from previous work in both scope and methodology, using a probe to measure similarities with continuous, visual representations. Closer to our goal of better understanding grounding is the work of \citet{cao2020behind}, that design probes for examining multi-modal models. In contrast, our work examines text-only models and does not rely on their ability to process images.

\paragraph{Language grounding.} A widely investigated research direction aims to connect natural language to the physical world \cite{bisk2020experience,mcclelland2019extending,tan-bansal-2019-lxmert,lu2019vilbert,lu201912,chen2019uniter,li2020oscar,tan2020vokenization}. This is typically done through training and evaluating models in tasks and datasets where both images and text are used, such as visual question answering (\citealp{antol2015vqa,hudson2019gqa}).
A number of previous work have investigated mappings between language and visual representations or mappings from both to a shared space. \citet{leong2011measuring} investigate semantic similarities between words and images through a joint latent space, finding a positive correlation with human rated similarities. Similarly, \citet{silberer-lapata-2014-learning} builds multi-modal representations by using stacked autoencoders. \citet{socher2013zero} and \citet{lazaridou-etal-2014-wampimuk} show that a shared latent space allows for zero-shot learning, demonstrating some generalization to previously unseen objects. \citet{lazaridou-etal-2015-combining} construct grounded word representations by exposing them to aligned visual features at training time. \citet{lucy-gauthier-2017-distributional} investigate how well word representations can predict perceptual and conceptual features, showing that a number of such features are not adequately predicted. \citet{collell2017imagined} uses word embeddings to create a mapping from language to visual features, using its outputs to build multimodal representations.
While our conclusions are generally aligned, our work differs from these in two important ways. Firstly, previous work studies context-independent word representations, while our method allows analysing language representations that depend on the context they are used in. We use this to examine a number of trained contextual language models. Secondly, while most previous work uses these mappings for building better grounded representations---often training the language models in the process---our work focuses on using them as a tool for inspecting already trained models, without modifying them. 

\paragraph{Zero-shot detection.} Recent work attempts to build object detectors that generalize to unseen object categories, by conditioning the predictions on word embeddings of the class \cite{rahman2018zero, demirel2018zero}, visual attributes \cite{demirel2018zero, zhu2019zero, mao2020zero} or text descriptions \cite{li2019zero}. In our work, we use language representations of words in context (captions) as inputs. More fundamentally, although our experiments on unseen object categories can be used for zero-shot detection, we differ from previous work in motivation, which translates to further experimental differences. Given our goal to analyse already trained models (as opposed to learning a generalizable object detector), we train nothing apart from a lightweight probe in our analyses.

\section{Probing contextual representations}
\label{sec:method}

\begin{figure*}
  \centering   
\includegraphics[width=.9\linewidth]{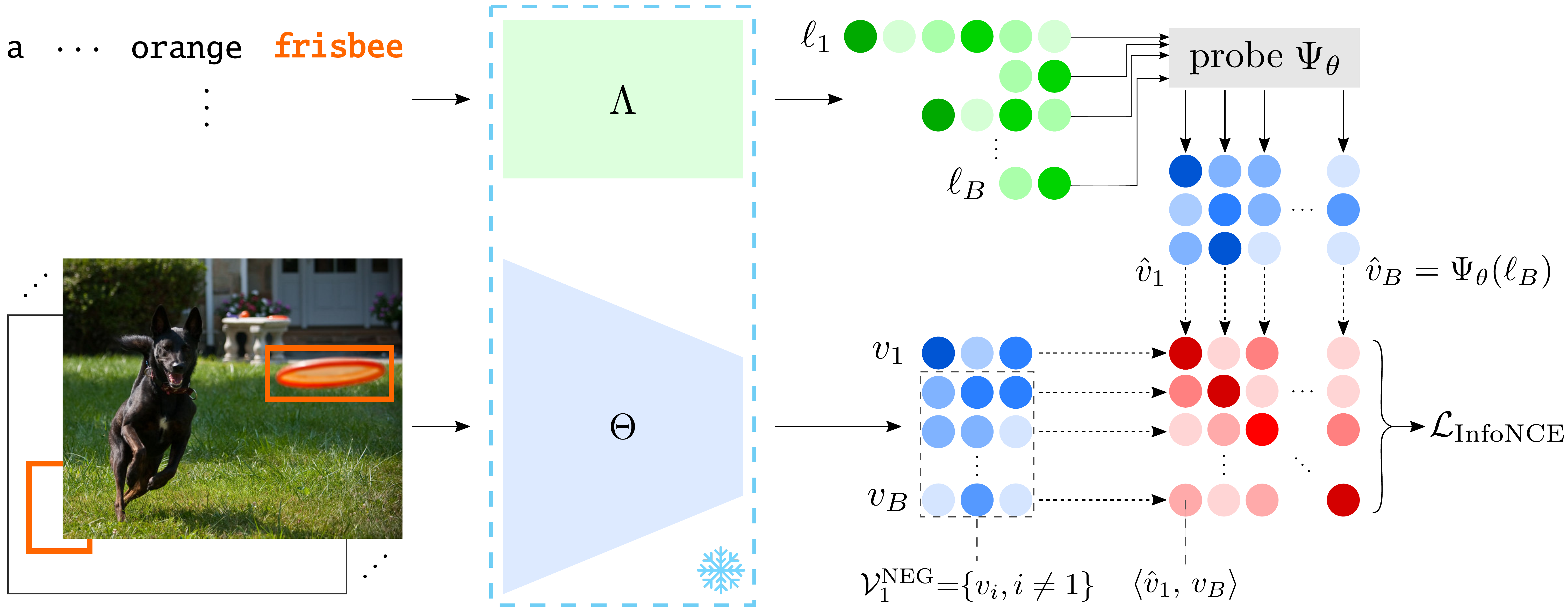}
  \caption{An overview of the proposed probing procedure. Frozen language and vision models ($\Lambda$ and $\Theta$) extract representations from matching pairs of words in text and objects in images. A probe $\Psi_\theta$ is trained to map representations from text (green) to visual (blue) domains while maximally preserving mutual information. For a given language representation $\ell_i$, the loss (Equation \ref{eq:infonce}) drives the probe's outputs $\hat{v}_i{=}\Psi_\theta(\ell_i)$ to be maximally useful for finding the aligned visual representation $v_i$ given all other visual representations in the batch ($\mathcal{V}^{\textrm{NEG}}_i={v_j, i\ne j}$). For such, only the pair-wise dot products $\langle\hat{v}_i,\, v_j\rangle$ are required (red).
  }
\label{fig:probe}
\end{figure*}

Our main goal is to characterize the relation between contextual language representations and the visual domain. 
We first describe how language and visual representations of concrete concepts can be collected from image captioning datasets (\cref{sec:representations}). Next, we design a probe that examines the relation between these representations, learning a mapping from language to visual representations (\cref{sec:probing}). An overview is illustrated in Figure \ref{fig:probe}. 

\subsection{Collecting data}
\label{sec:representations}

At the center of our analysis are contextual representations of visually observable nouns, which we refer to as \textit{object categories}.
Here, we describe how pairs of matching language and visual representations $(\ell,v)$ are collected from image captioning datasets.

\paragraph{Language representations $(\ell)$}  are extracted from image captions. To accommodate recent language models and tokenizers, we allow such representations to be contextual and have variable length,\footnote{Conforming with sub-word tokenizers or multi-word expressions such as \txtstyless{\objstyle{fire extinguisher}}.} where each element in $\ell$ has a fixed dimension $d_L$. The length of the representations $\ell$ for each object category is determined by the tokenizer. We treat a model that extracts representations from text as a function $\Lambda$ that maps a string $o$ (here, object categories) in a larger textual context $c$ (here, captions) to the representation $\ell  =  \Lambda(o \mid c)$.
This formalism also encompasses non-contextual embeddings, with $ \Lambda(o \mid c)=\Lambda(o)$.

\paragraph{Visual representations $(v)$}  are extracted from objects in images using a trained object detection model $\Theta$. For simplicity, we use $v = \Theta(o \mid i)$ to refer to the extracted features corresponding to the detected object from image $i$ that is both 1) classified as a member of object category $o$ and 2) assigned the highest confidence by the model among those. Visual representations $\Theta(o \mid i)$ have fixed dimensions $d_V$.

\paragraph{Paired data $(\ell, v)$} with aligned representations is collected from an image captioning dataset with paired captions $c$ and images $i$. For each image $i$, and each object $o$ detected by the object detector $\Theta$, if $o$ appears in some associated caption $c$, we include the pair $(\ell = \Lambda(o \mid c), v = \Theta(o \mid i))$. To avoid having multiple pairs $(\ell,v)$ associated with the same visual instance, we ensure that at most one pair $(\ell,v)$ per object category in each image is included. In this work, we use the 1600 object categories from Faster R-CNN \cite{ren2015faster} trained on Visual Genome \cite{krishna2017visual}.

\subsection{Probing representations}
\label{sec:probing}

At a high level, language representations are inspected via a shallow neural probing model (Figure \ref{fig:probe}). In training, the probe learns a mapping from language to visual representations (\cref{subsec:training}). We then evaluate the quality of these mappings by measuring how well they can be used to retrieve matching image patches (\cref{subsec:eval}).

\subsubsection{Training the probe}
\label{subsec:training}

The probe is optimized to maximally preserve the mutual information between the distributions of language and visual representations. This is done via InfoNCE \cite{oord2018representation} (Equation \ref{eq:infonce}), a loss function commonly used for retrieval and contrastive learning \cite{le2020contrastive}. We note the mutual information is a bottleneck on how well two random variables can be mapped to one another, given its relation to conditional entropy. In training, the probe $\Psi_\theta$ with parameters $\theta$ takes inputs $\ell$ and estimates visual representations $\hat{v} = \Psi_\theta(\ell)$ with the same dimensionality $d_V$ as the corresponding visual representations $v$. For each pair $(\ell, v)$, this loss relies on a set of distractors $\mathcal{V}^{\textrm{NEG}}_\ell$, containing visual representations which are \textit{not} aligned with the language representations $\ell$. The representations in $\mathcal{V}^{\textrm{NEG}}_\ell$ are used for contrastive learning and are drawn from the same visual model, using different objects or images. Minimizing this loss drives the dot product $\langle\Psi_\theta(\ell)\,, u\rangle$ to be maximal for $u = v$ and small for all $u \in \mathcal{V}^{\textrm{NEG}}_\ell$. In other words, training pushes the estimates $\hat{v} = \Psi_\theta(\ell)$ to be maximally useful in discerning between positive and negative visual pairings.
\begin{equation}
    \mathcal{L}_{\textrm{InfoNCE}} = -\mathbb{E}_\ell\left[\log\frac{e^{\langle \Psi_\theta(\ell)\,, v_\ell\rangle}}{\sum\limits_{v' \in \{v\}\bigcup \mathcal{V}^{\textrm{NEG}}_{\ell}} e^{\langle \Psi_\theta(\ell)\,, v'\rangle}}\right]
    \label{eq:infonce}
\end{equation}

In practice, the expectation in Equation \ref{eq:infonce} is estimated over a batch of size $B$ with samples of aligned language and visual representations $((\ell_1, v_1), \dots, (\ell_B, v_B))$. For efficiency, we use other visual representations in the batch as distractors for a given representation ($\mathcal{V}^{\textrm{NEG}}_i = \{v_j, j \ne i\}$). Thus, only the dot products $\langle\hat{v}_i= \Psi_\theta(\ell_i), \, v_j\rangle$ are needed to calculate the loss, as illustrated in Figure \ref{fig:probe}.  Importantly, we note that the models used to extract representations are not trained or changed in any way during the probing procedure.

\subsubsection{Evaluation procedure}
\label{subsec:eval}

For evaluation, we compute recall in retrieving image patches given objects in text, using new pairs of language and visual representations from unseen images and captions. Consider the set of all collected visual representations for evaluation, $\mathcal{V}$. For each language representation $\ell$, we use the trained probe to generate our estimate $\hat{v} = \Psi_\theta(\ell)$, and find the instances $v' \in \mathcal{V}$ that maximize the dot product $\langle \hat{v}\,, v'\rangle$. Given an integer $k$, we consider recall at $k$ at both instance and category levels. Formally: 

\paragraph{Instance Recall (IR@k)} measures how frequently the correct visual instance is retrieved. More precisely, it is the fraction of pairs $(\ell,v)$ where the instance $v$ is in the top-$k$ visual representations retrieved from $\hat{v} = \Psi_\theta(\ell)$.
\paragraph{Category Recall (CR@k)} measures how frequently instances of the correct object category are retrieved. More precisely, it is the fraction of pairs $(\ell,v=\Theta(o \mid i))$ where any of the top-$k$ retrieved visual representations $v'=\Theta(o' \mid i')$ belongs to the same object category as $v$ (i.e. $o' = o$).

\paragraph{}Higher IR and CR scores indicate better performance and, by definition, CR@k cannot be smaller than IR@k. These metrics form the basis of our evaluation, and we take multiple steps to promote experimental integrity. Learned mappings are evaluated in two scenarios, where pairs $(\ell,v)$ are collected using object categories either \textit{seen} or \textit{unseen} by the probe during training. The later is the focus of the majority of our experiments. For both scenarios, images and captions have no intersection with those used in training. Further, we create multiple \textit{seen}/\textit{unseen} splits from our data, training and testing on each split. We then report average and standard deviation of the recall scores across 5 splits.

\section{Experimental settings}
\label{sec:experiments}

\subsection{Language models}
The majority of examined models are contextual representation models based on the transformer architecture \cite{vaswani2017attention} trained on text-only data.
We examine the \textit{base} ($d_L = 768$) and \textit{large} ($d_L = 1024$) versions of BERT uncased, RoBERTa, ALBERT and T5 \cite{devlin-etal-2019-bert, liu2019roberta, lan2019albert, 2019t5}. For T5, we also examine the \textit{small} version, with $d_L = 512$. For all these models, we use pre-trained weights from the HuggingFace Transformers library \cite{Wolf2019HuggingFacesTS}\footnote{\href{https://github.com/huggingface/transformers}{https://github.com/huggingface/transformers}}, and use representations from the last layer. Additionally, we inspect non-contextual representations using GloVe embeddings \cite{pennington2014glove}, using embeddings trained on 840 billion tokens of web data, with $d_L = 300$ and a vocabulary size of 2.2 million.\footnote{\href{https://nlp.stanford.edu/projects/glove/}{https://nlp.stanford.edu/projects/glove/}}

\subsection{Vision models} As is common practice in natural language grounding literature \cite{anderson2018bottom, tan-bansal-2019-lxmert, su2019vl, lu201912}, we use a Faster R-CNN model \cite{ren2015faster} trained on Visual Genome \cite{krishna2017visual} to extract visual features with $d_V=2048$. We use the trained network provided by \citet{anderson2018bottom}\footnote{\href{https://github.com/peteanderson80/bottom-up-attention}{https://github.com/peteanderson80/bottom-up-attention}}, and do not fine-tune during probe training.

\subsection{Data}
We collect representations from two image captioning datasets, Flickr30k \cite{young2014image}, with over 150 thousand captions and 30 thousand images, and MS-COCO \cite{lin2014microsoft}, with 600 thousand captions and 120 thousand images in English. The larger MS-COCO is the focus of the majority of our experiments. We build \textit{disjoint} training, validation and test sets from the aggregated training and validation image captions. To examine generalization to new objects, we test on representations from both \textit{seen} or \textit{unseen} object categories, built from images and captions not present in the training data. From the 1600 object categories of our object detector, we use 1400 chosen at random for training and \textit{seen} evaluation. The remaining 200 are reserved for \textit{unseen} evaluation. Furthermore, we train and test our probe 5 times, each with a different 1400/200 split of the object categories. For each object category split, we build validation and test sets with sizes proportional to the number of object categories present: \textit{seen} test sets contain 7000 representation pairs and \textit{unseen} test sets contain 1000 pairs. The validation sets used for development consists of \textit{seen} object categories, with the same size as the \textit{seen} test sets. All remaining data is used for training.

\subsection{Control task}

Contrasting the probe performance with a control task is central to probing \cite{hewitt2019control}. We follow this practice by learning in a control task where representations are mapped to \textit{permuted} visual representations. More precisely, we replace each visual representation $v = \Theta(o \mid i)$ with another $v' = \Theta(o' \mid i')$ chosen at random from an object category $o'=f(o)$ that depends on the original object category $o$. Here, $f$ dictates a random permutation of the object categories. For instance, visual representations of the original category \txtstyle{\objstyle{cat}} are replaced with representations from a second category \txtstyle{\objstyle{dog}}; representations from the category \txtstyle{\objstyle{dog}} are replaced by those from \txtstyle{\objstyle{tree}}, and so on.

\subsection{Implementation and hyper-parameters}

Our probe consists of a shallow neural model. To process the naturally sequential language representations $\ell$, we use a single-layered model with LSTM cells \cite{hochreiter1997long} with 256 hidden units and only unidirectional connections. The outputs are then projected by a linear layer to the visual space. The probe is trained using Adam optimizer \cite{kingma2014adam} with a learning rate of 0.0005, weight decay of 0.0005 and default remaining coefficients ($\beta_1{=}0.9$ $\beta_2{=}0.999$ and $\epsilon{=}10^{-8}$). We train with a batch size of 3072, for a total of 5 epochs on one GPU.

\section{Results and discussion}
\label{sec:results}

At a high level, our experiments show that i) language representations are strong signals for choosing between different visual features both at the instance and category levels; ii) context is largely helpful for instance retrieval; iii) InfoNCE works better than other studied losses, and some consistency is found across datasets; iv) visually grounded models outperform text-only models; v) all models lag greatly behind human performance. We provide further details in  \S\ref{sec:retrieval}-\ref{sec:analyses}.

\begin{table}
\small
\renewcommand{\arraystretch}{1.15}
\setlength\tabcolsep{3.2pt}
    \centering
    \begin{tabular}{rl@{\hskip .1in}c@{\hskip .1in}c@{\hskip .1in}c} \Xhline{2\arrayrulewidth}
    \textbf{\#} & \textbf{Experiment} & \textbf{IR@1} & \textbf{IR@5} & \textbf{CR@1}
    \\\Xhline{2\arrayrulewidth}
0 & Random & 0.1 $\pm$ 0.1 & 0.5 $\pm$ 0.2 & 6.0 $\pm$ 2.0 \\\
1 & Control & 0.0 $\pm$ 0.0 & 0.4 $\pm$ 0.2 & 3.0 $\pm$ 1.4 \\\hline
2 & GloVe & 5.1 $\pm$ 0.5 & 18.5 $\pm$ 1.4 & 87.3 $\pm$ 3.5\\\
3 & \underline{BERT base} & 12.0 $\pm$ 1.0 & 36.0 $\pm$ 0.9 & 88.1 $\pm$ 2.4\\\
4 & BERT large & 11.6 $\pm$ 0.7 & 34.9 $\pm$ 2.6 & 89.3 $\pm$ 2.4\\\
5 & RoBERTa base & 11.6 $\pm$ 0.3 & 34.4 $\pm$ 2.2 & 90.4 $\pm$ 0.6\\\
6 & RoBERTa large & 10.9 $\pm$ 1.1 & 32.8 $\pm$ 2.5 & 88.7 $\pm$ 3.2\\\
7 & ALBERT base & 8.7 $\pm$ 0.2 & 28.8 $\pm$ 1.6 & 84.4 $\pm$ 2.1\\\
8 & ALBERT large & 9.4 $\pm$ 1.0 & 28.8 $\pm$ 2.3 & 84.2 $\pm$ 4.2 \\\
9 & T5 small & 10.1 $\pm$ 0.7 & 32.9 $\pm$ 1.5 & 87.2 $\pm$ 4.1 \\\
10 & T5 base & 10.8 $\pm$ 0.8 & 33.3 $\pm$ 2.3 & 85.3 $\pm$ 2.8 \\\
11 & T5 large & 11.8 $\pm$ 0.5 & 34.7 $\pm$ 2.1 & 87.2 $\pm$ 2.4
 \\\Xhline{2\arrayrulewidth}
    \end{tabular}
    \caption{Average instance recall (IR@k) and category recall (CR@k) for test sets with \textit{unseen} object categories. For each model, we train and evaluate 5 times, using different sets of object categories seen in training. Unlike the control task with permuted representations, mappings learned from sensible representations generalize well to unseen object categories.}
    \label{tab:main_unseen}
\end{table}

\subsection{Retrieval results} 
\label{sec:retrieval}
Table \ref{tab:main_unseen} summarizes instance and category retrieval performance for different language models and control experiments, using test data with \textit{unseen} object categories. Our results indicate that language representations alone are strong signals for predicting visual features: for all examined language models, recall scores are significantly better than random and control. Qualitative results can be found in Figure \ref{fig:examples}. We note that category recall scores are significantly higher than instance recall. This is reasonable since there are many more positive alignments at the category level.
Compared to other inspected models, BERT base shows the best results for instance retrieval, and will be the focus of further analyses.

Contrasting the performance of non-contextual representations from GloVe with that of contextual models shows that context considerably affects instance recall. For instance, GloVe and BERT base yield 5.1\% to 12.0\% IR@1, respectively. This gap is sensible, since a non-contextual representation should not be able to discern between distinct image patches depicting the same object category. While still lagging behind a number of contextual representations, we observe strong category recall for GloVe, which we hypothesize is due to the ease in predicting the correct output category since input representations are fixed, independently of context. We further explore the role of context in \cref{sec:analyses}.

\begin{table}
\small
\renewcommand{\arraystretch}{1.15}
\setlength\tabcolsep{4.2pt}
    \centering
    \begin{tabular}{rl@{\hskip .1in}c@{\hskip .1in}c@{\hskip .1in}c} \Xhline{2\arrayrulewidth}
    \textbf{\#} & \textbf{Experiment} & \textbf{IR@1} & \textbf{IR@5} & \textbf{CR@1}
    \\\Xhline{2\arrayrulewidth}
    
0 & Random & 0.1 $\pm$ 0.1 & 0.1 $\pm$ 0.1 & 1.2 $\pm$ 0.1 \\
1 & Control & 1.6 $\pm$ 0.1 & 7.8 $\pm$ 0.6 & 41.3 $\pm$ 5.6 \\
2 & BERT base & 14.9 $\pm$ 0.3 & 43.4 $\pm$ 0.8 & 90.4 $\pm$ 0.4 
 \\\Xhline{2\arrayrulewidth}
    \end{tabular}
    \caption{Average instance recall (IR@k) and category recall (CR@k) for test sets with \textit{seen} object categories.}
    \label{tab:main_seen}
\end{table}

Moreover, Table \ref{tab:main_seen} shows performance on test sets with \textit{seen} object categories. Comparing with Table \ref{tab:main_unseen}, BERT representations show good generalization to unseen object categories. This generalization is consistent with previous observations on zero-shot experiments, using non-contextual word embeddings \cite{lazaridou-etal-2014-wampimuk}.

Finally, our results attest to the selectivity of the probe: for the control task with permuted representations (Tables \ref{tab:main_unseen} and \ref{tab:main_seen}, Row 1), substantially lower performance is found. This gap is particularly high for \textit{unseen} object categories, where only sensibly paired representations perform better than chance.

\begin{table}
\small
\renewcommand{\arraystretch}{1.15}
\setlength\tabcolsep{4.6pt}
    \centering
    \begin{tabular}{l@{\hskip .2in}c@{\hskip .1in}c@{\hskip .1in}c} \Xhline{2\arrayrulewidth}
    \textbf{Loss function} & \textbf{IR@1} & \textbf{IR@5} & \textbf{CR@1}
    \\\Xhline{2\arrayrulewidth}
MSE & 3.0 $\pm$ 0.3 & 12.1 $\pm$ 1.3 & 57.5 $\pm$ 8.7 \\
Neg. cosine sim. & 6.9 $\pm$ 0.7 & 23.4 $\pm$ 1.3 & 75.1 $\pm$ 6.3\\

Triplet loss& 8.4 $\pm$ 0.6 & 28.8 $\pm$ 0.9 & 81.7 $\pm$ 3.6 \\
\underline{InfoNCE} &  12.0 $\pm$ 1.0 & 36.0 $\pm$ 0.9 & 88.1 $\pm$ 2.4
 \\\Xhline{2\arrayrulewidth}
    \end{tabular}
    \caption{Comparison in retrieval performance on \textit{unseen} object categories for different training losses, using representations from BERT base. InfoNCE yields better results than other loss functions.}
    \label{tab:losses}
\end{table}

\subsection{Ablations}
\label{sec:ablations}

\paragraph{Loss ablations.} In addition to InfoNCE, we ablate on 3 other loss functions: mean squared error (MSE), negative cosine similarity, and triplet loss\footnote{$\mathcal{L}_{trip} =
    \mathbb{E}_\ell
    [\max(\delta_{\ell,v'_\ell} - \delta_{\ell,v_\ell} + \alpha, 0)]$, where the margin $\alpha$ is  set to 1.0, $v'\in \mathcal{V}^{\textrm{NEG}}$ and $\delta_{\ell,v} =
\cos({\Psi_\theta(\ell), v_\ell})$.}. 
The results for unseen object categories are summarized in Table \ref{tab:losses}: while all losses yield better than random results, InfoNCE performs the best. This validates the theoretical intuition that InfoNCE would be advantageous, as it allows for directly optimizing the probe to maximally preserve the mutual information between the representations, a bottleneck on the remaining entropy after the mapping. 

\paragraph{Data ablations.} In addition to MS-COCO, which is the used for the majority of our experiments, we show results with data collected from the smaller Flickr30k. We report the test retrieval performance for unseen object categories using representations from BERT base in Table \ref{tab:data_ablations}. These results indicate consistency across the datasets, despite their considerable difference in size.
\begin{table}
\small
\renewcommand{\arraystretch}{1.15}
\setlength\tabcolsep{3.1pt}
    \centering
    \begin{tabular}{lc@{\hskip .05in}c@{\hskip .05in}c} \Xhline{2\arrayrulewidth}
    \textbf{Dataset} & \textbf{\# Images / \# Captions} & \textbf{IR@1} & \textbf{CR@1}
    \\\Xhline{2\arrayrulewidth}
    MS-COCO & 120k / 600k & 12.0 $\pm$ 1.0 & 88.1 $\pm$ 2.4 \\
    Flickr30k & 30k / 150k & 9.8 $\pm$ 0.9 & 85.6 $\pm$ 3.4 \\
    \Xhline{2\arrayrulewidth}
    \end{tabular}
    \caption{Comparison for different datasets in retrieval performance of \textit{unseen} object categories with representations from BERT base. Despite large differences in size, results indicate consistency across datasets.}
    \label{tab:data_ablations}
\end{table}

\subsection{Analyses}
\label{sec:analyses}

\begin{figure}
  \centering   
\includegraphics[width=.9\linewidth]{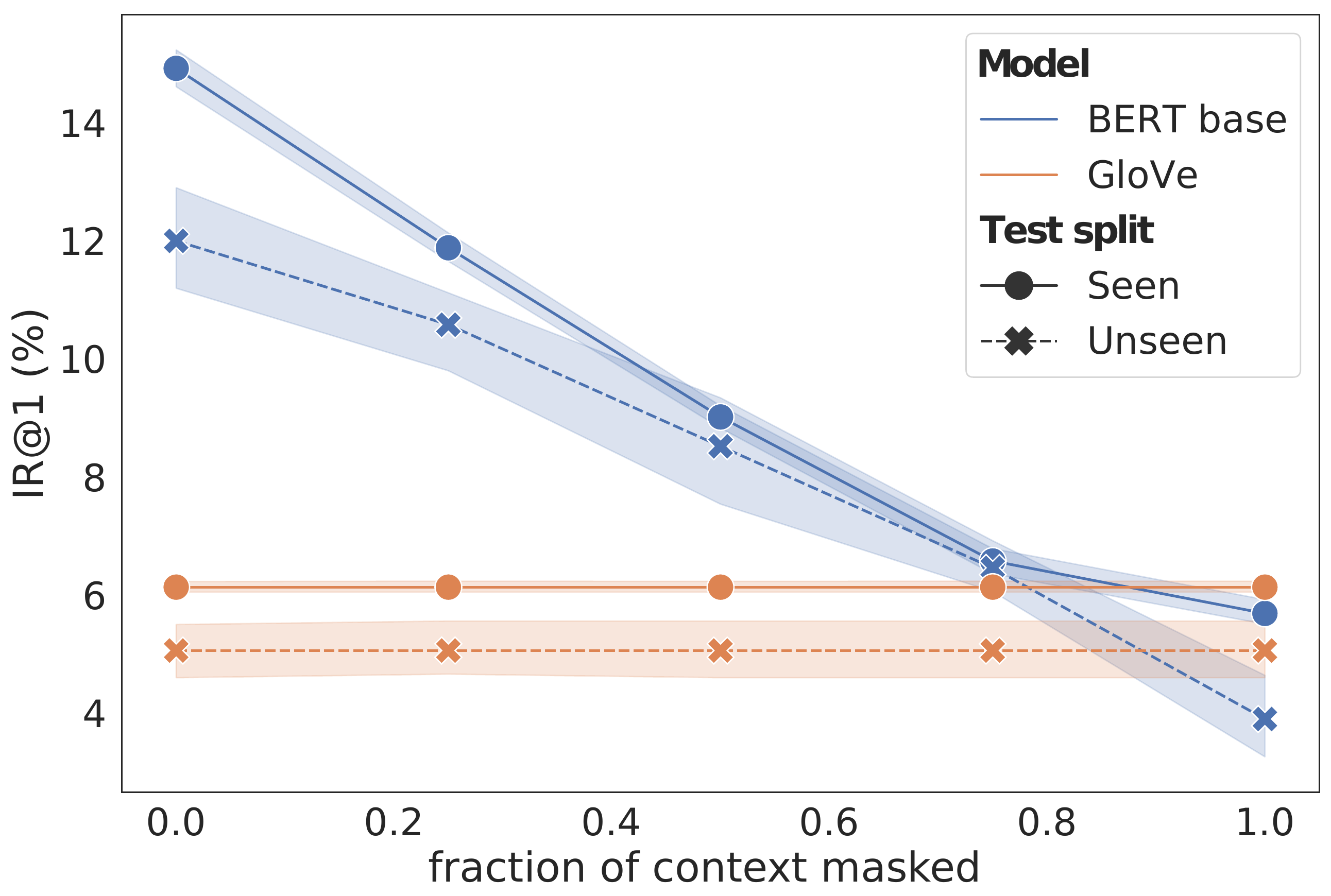}
  \caption{Instance recall as context tokens are progressively masked out. Retrieval performance for BERT quickly degrades as higher proportions of the context are masked.}
\label{fig:masked_context}
\end{figure}

\paragraph{Influence of context.} 
 We study whether the gap in instance retrieval performance from GloVe and BERT comes from the use of context or intrinsic differences of these models. This is explored by measuring how instance recall varies as we probabilistically mask out context tokens in the captions at different rates. As shown in Figure \ref{fig:masked_context}, performance drops substantially as more tokens are masked; in the limit where only the object tokens remain (i.e. the fraction of context masked is 1.0), BERT's representations perform marginally worse than the non-contextual GloVe embeddings.

\begin{figure}
  \centering   
\includegraphics[width=.85\linewidth]{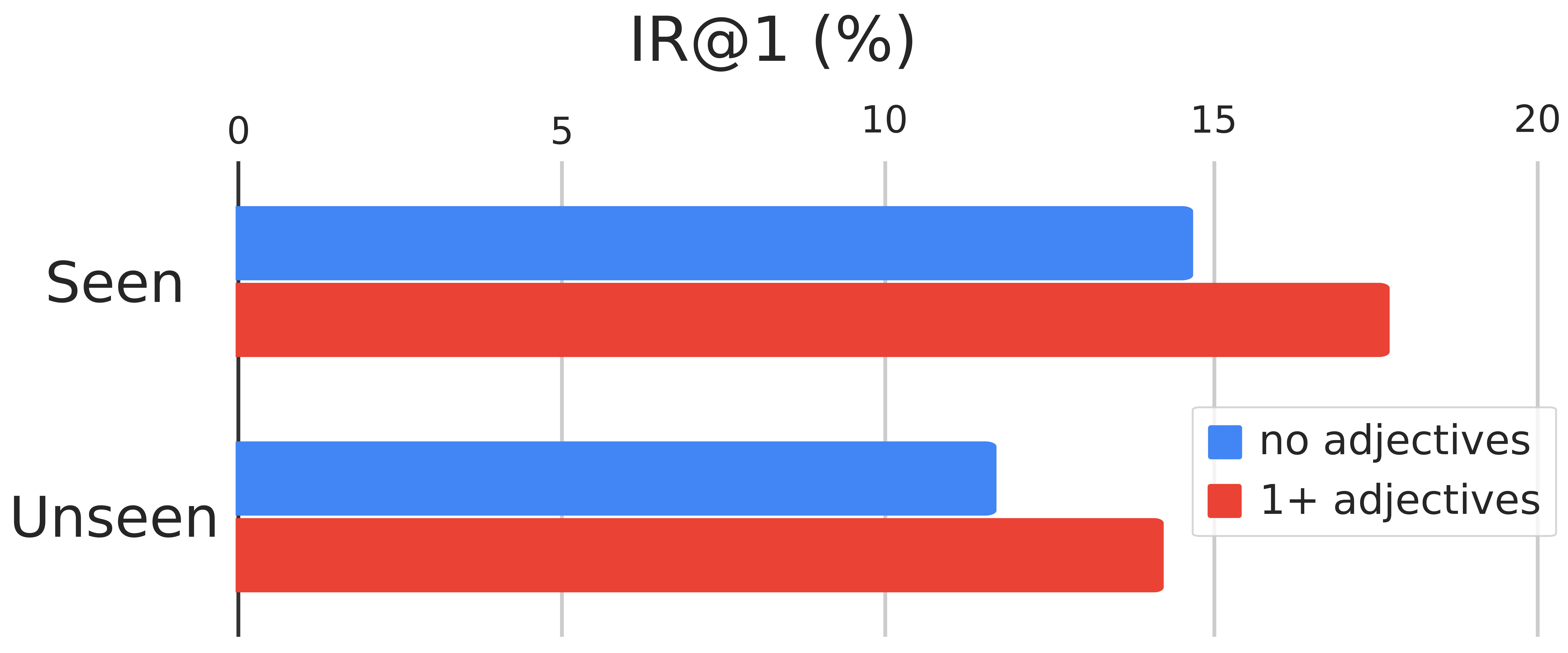}
  \caption{More descriptive contexts enable more accurate retrieval. In the plot, we show instance recall at 1 when object categories are or are not accompanied by adjectives, using representations from BERT base.}
\label{fig:adjectives}
\end{figure}

Figure \ref{fig:adjectives} compares instance-level retrieval accuracy for representations when objects  have none or at least one adjective associated with them, as processed by the dependency parser from AllenNLP library \cite{gardner2018allennlp}. These adjectives commonly include colors (e.g. \txtstyle{white}, \txtstyle{black}) and sizes (e.g. \txtstyle{big}, \txtstyle{small}), indicating contextual information. The results show clear gains in instance recall when objects are accompanied by adjectives, confirming that context enables more accurate retrieval. We refer back to Figure \ref{fig:examples} for qualitative results on the influence of context.

\paragraph{Grounded language models.} We further inspect representations from several grounded language models, namely LXMERT, VL-BERT (base and large) and VILBERT-MT \cite{tan-bansal-2019-lxmert,su2019vl,lu2019vilbert,lu201912}). While these models typically process visual and textual inputs jointly, we adapt them to include only the language branches, restricting attention to the text inputs.
For all these models, we use the code and weights made public by the authors.\footnote{ \href{https://github.com/airsplay/lxmert}{github.com/airsplay/lxmert}; \href{https://github.com/jackroos/VL-BERT}{github.com/jackroos/VL-BERT}; \href{https://github.com/facebookresearch/vilbert-multi-task}{github.com/facebookresearch/vilbert-multi-task}}
The results, summarized in Table \ref{tab:grounded}, show that grounded models slightly outperform the ungrounded BERT base. At the category level, we see small relative differences in performance between grounded and ungrounded models. At the instance level, the relative improvement is higher, especially for VILBERT-MT, while still much lower than human performance as shown in the next experiment. 

\begin{table}
\small
\renewcommand{\arraystretch}{1.15}
\setlength\tabcolsep{4.6pt}
    \centering
    \begin{tabular}{l@{\hskip .2in}c@{\hskip .1in}c@{\hskip .1in}c} \Xhline{2\arrayrulewidth}
    \textbf{Model} & \textbf{IR@1} & \textbf{IR@5} & \textbf{CR@1}
    \\\Xhline{2\arrayrulewidth}
BERT base & 12.0 $\pm$ 1.0 & 36.0 $\pm$ 0.9 & 88.1 $\pm$ 2.4\\\hline
LXMERT & 13.7 $\pm$ 1.0 & 39.2 $\pm$ 2.5 & 90.3 $\pm$ 1.2\\
VL-BERT base & 12.5 $\pm$ 1.0 & 37.6 $\pm$ 1.1 & 88.7 $\pm$ 1.4\\
VL-BERT large &  12.6 $\pm$ 1.1 & 37.5 $\pm$ 2.4 & 88.7 $\pm$ 2.3 \\
VILBERT-MT & 15.4 $\pm$ 1.2 & 42.4 $\pm$ 2.7 & 90.8 $\pm$ 1.9
 \\\Xhline{2\arrayrulewidth}
    \end{tabular}
    \caption{Retrieval performance for \textit{unseen} object categories, using representations from BERT and a number of grounded language models.}
    \label{tab:grounded}
\end{table}

\paragraph{Human performance.}
Finally, we contrast the examined models with human performance in  retrieving visual patches given words in sentences. Such a comparison helps disentangling the quality of the learned mappings with possible incidental matches, i.e., language representations with more than one positive visual match. As they are also affected by these artifacts, human subjects offer a sensible point of comparison. In virtue of the limited human attention, we evaluate on a reduced test set with unseen object categories, randomly sampling 100 data points from it. For each object in a sentence, subjects are presented with 100 image patches and asked to choose the closest match. We collect over 1000 annotations from 17 in-house annotators, with at least 30 annotations each. Our results are shown in Table \ref{tab:human}. On the same test set, we find a large gap from learned mappings for both grounded and ungrounded models to human performance, exposing much room for  improvement.

\begin{table}
\setlength\tabcolsep{4.2pt}
\renewcommand{\arraystretch}{1.08}
\small
    \centering
    \begin{tabular}{cccc} \Xhline{2\arrayrulewidth}
      \textbf{Chance} & \textbf{BERT base} & \textbf{VILBERT-MT} & \textbf{Human} \\\Xhline{2\arrayrulewidth}
    1\% & 43\% & 53\% & 76\% \\\Xhline{2\arrayrulewidth}
\end{tabular}
    \caption{A sizable gap in instance recall (IR@1) is seen by comparing the performance of humans and the examined models in a reduced test set with 100 samples.}
    \label{tab:human}
\end{table}

\section{Conclusion}
\label{sec:conclusion}

Understanding the similarities between language and visual representations has important implications on the models, training paradigms and benchmarks we design. We introduced a method for empirically measuring the relation between contextual language representations and corresponding visual features. We found contextual  language models to be useful---while far from human subjects---in discerning between different visual representations. Moreover, we explored how these results are influenced by context, loss functions, datasets and explicit grounding during training. Altogether, we hope that our new methodological and practical insights foster further research in both understanding the natural connections between language and visual representations and designing more effective models at the intersection the two modalities. 

\section*{Acknowledgements}
This research was supported by the grants from ONR N00014-18-1-2826, DARPA N66001-19-2-4031, 67102239, NSF III-1703166, IIS-1652052, IIS-17303166, and an Allen Distinguished Investigator Award and a Sloan Fellowship. Authors would also like to thank Raymond J. Mooney and members of the UW-NLP, H2Lab and RAIVN Lab at the University of Washington for their valuable feedback and comments.

\bibliography{main}
\bibliographystyle{acl_natbib}

\end{document}